\documentclass[runningheads]{llncs}
\usepackage[T1]{fontenc}
\usepackage{accv}
\usepackage{accvabbrv}

\usepackage{xcolor}
\usepackage{graphicx}
\usepackage{booktabs}
\usepackage{multirow}
\usepackage{caption}
\usepackage{ulem}
\usepackage{makecell}
\usepackage{hyperref}
\usepackage[accsupp]{axessibility}
\begin{document}
\title{Enhanced Super-Resolution Training via Mimicked Alignment for Real-World Scenes}
\titlerunning{Mimicked Alignment for Real-World SR}
%
\author{Omar Elezabi\and
Zongwei Wu\thanks{Corresponding Author}\and
Radu Timofte}
\authorrunning{O. Elezabi et al.}
%
\institute{Computer Vision Lab, CAIDAS \& IFI, University of Würzburg
\email{\{omar.elezabi,zongwei.wu,radu.timofte\}@uni-wuerzburg.de}}
\maketitle      
\begin{abstract}
Image super-resolution methods have made significant strides with deep learning techniques and ample training data. However, they face challenges due to inherent misalignment between low-resolution (LR) and high-resolution (HR) pairs in real-world datasets. In this study, we propose a novel plug-and-play module designed to mitigate these misalignment issues by aligning LR inputs with HR images during training. Specifically, our approach involves mimicking a novel LR sample that aligns with HR while preserving the degradation characteristics of the original LR samples. This module seamlessly integrates with any SR model, enhancing robustness against misalignment. Importantly, it can be easily removed during inference, therefore without introducing any parameters on the conventional SR models. We comprehensively evaluate our method on synthetic and real-world datasets, demonstrating its effectiveness across a spectrum of SR models, including traditional CNNs and state-of-the-art Transformers. The source codes will be publicly made available at \href{https://github.com/omarAlezaby/Mimicked_Ali}{github.com/omarAlezaby/Mimicked\_Ali}

\end{abstract}
%
%
%

\section{Introduction}
\label{sec:intro}

\begin{figure}[tb]
  \centering
    \includegraphics[trim={0 0 0 1mm},clip,width=\textwidth]{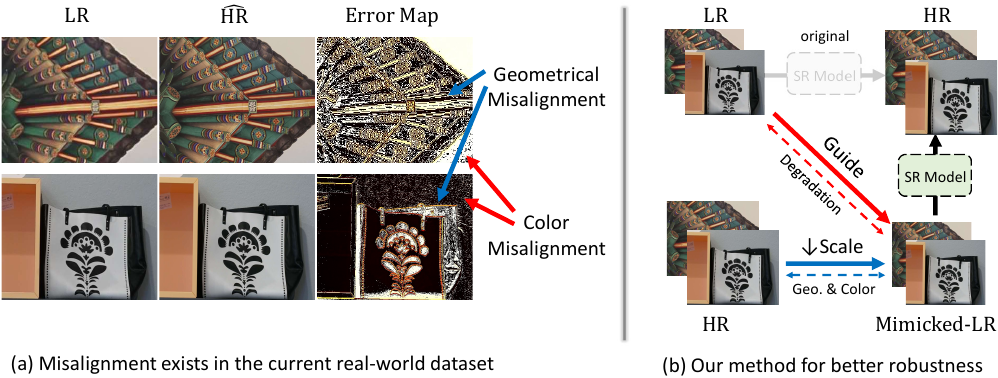}
  \caption{\textbf{Motivation:} Despite significant effort, misalignment issues persist in real-world datasets \cite{cai2019toward,zhang2019zoom}, limiting the potential of SR models. To address this challenge, we propose a novel alignment method using mimicked-LR, which maintains the same geometry and color properties as the HR input while sharing the same degradation type as the LR. By training our network with this newly generated mimicked-LR, we fully leverage the potential of SR models. $\widehat{HR}$ is the downscaled HR.}
  \label{fig:miss-alignment}
\end{figure}

\begin{figure}[tb]
  \centering
    \includegraphics[trim={0 1.5mm 0 3.5mm},clip,width=\textwidth]{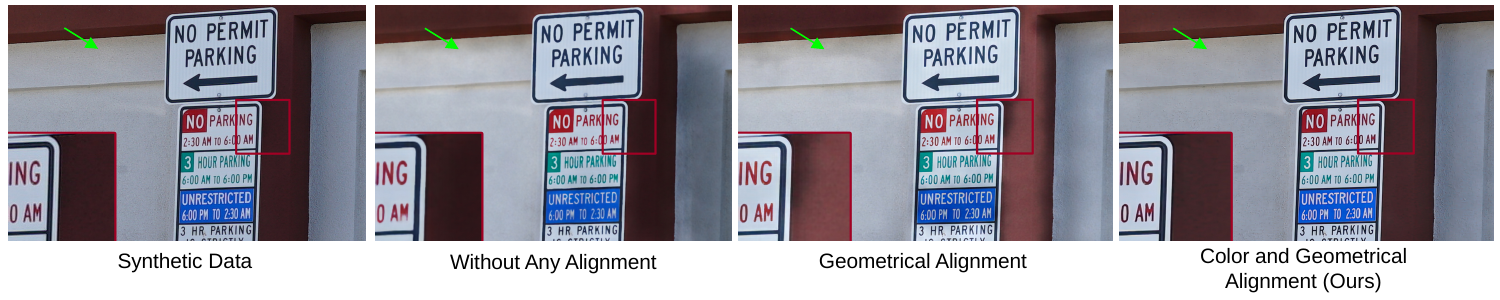}
  \caption{The importance of Color and Geometrical Alignment between the LR and HR images for the training on Realistic SR Datasets}
  \label{fig:teaser}
\end{figure}

Single image super-resolution (SISR) aims to recover fine details from a low-resolution (LR) image while upscaling it. This process is crucial for enhancing image quality, particularly with the surge of streaming services and augmented reality (AR) applications \cite{khani2021efficient,zhang2021benchmarking}. With the development of deep learning methods \cite{yang2019deep,zhou2023srformer,liu2020residual,sun2022shufflemixer} and the availability of large-scale data \cite{nah2017deep,abdelhamed2018high}, recent works have shown great success in improving the image quality \cite{liang2021swinir,wang2023ultra}.

The effectiveness of learning models is highly dependent on the quantity and quality of LR-HR (high-resolution) image pairs. Synthetic data \cite{agustsson2017ntire,lim2017enhanced} has been a popular choice due to its ability to provide ample high-quality training pairs. However, synthetic data often oversimplify the SR problem by using uniform degradations like bicubic or Gaussian blurring \cite{agustsson2017ntire,timofte2017ntire}. Consequently, models trained on such data struggle to generalize to real-world scenarios with complex degradations \cite{kohler2019toward}. As shown in Fig. \ref{fig:teaser}, models trained on synthetic datasets produce less sharp images with noise when applied to real data. To address this, some previous works have proposed more complex degradation pipelines to create better-simulated datasets \cite{bell2019blind,zhang2021designing}. However, these efforts often fail to capture specific system degradations, such as those introduced by different lenses.

A more realistic approach involves creating real-world super-resolution datasets using DSLR cameras with various zoom lenses \cite{cai2019toward,zhang2019zoom}. Despite producing images with real degradations, these datasets often suffer from misalignment issues between LR and HR images. Practical systems, when adjusting lens settings, introduce significant geometric misalignments, such as translations, scaling, and rotation. Additionally, changes in optical settings result in varying light readings for the same scene, affecting pixel colors, brightness, and illumination. These misalignments significantly limit the effectiveness of existing SR models.

In this work, we thoroughly analyze the existing alignment efforts with the goal of developing a more generic, holistic, and easy-to-implement method to improve the robustness of SR models. We study the problem under more rigorous settings. We propose training existing models on misaligned LR-HR data and evaluating their performance on perfectly aligned synthetic datasets to focus solely on the detail-recovery capabilities of SR models. Current approaches involve creating alternative inputs that better match the HR images \cite{zhang2022self,sun2021learning}. For example, Zhang \etal \cite{zhang2022self} uses the center crop of short-focus and telephoto images to construct LR-HR pairs. However, this approach is limited to small and simple shifts and fails with more complex misalignments. Sun \etal \cite{sun2021learning} leverages unpaired images to learn real-world degradations and generate alternative inputs, but this method does not fully utilize paired LR-HR data and complete supervision. Regarding color mismatch, few works have addressed this issue \cite{chen2022real,zhang2021learning}. A related work for under-display cameras \cite{feng2023generating} introduces a domain adaptation module to transfer color information between a high-quality reference and a degraded image. In addition to another module for geometrical alignment. However, these modules are trained separately which results in error accumulation that produces unsatisfactory visual results for our SR problem.

To address these issues, we propose a modification module to create a more realistic and aligned LR image, denoted as LR Mimicking Module. Our method takes the LR ($LR$) and downscaled HR ($\widehat{HR}$) as input and aims to transform the downscaled HR into the Mimicked-LR ($Mim_{LR}$) by incorporating characteristics from the LR input. This approach ensures that the produced $Mim_{LR}$ is geometrically and color-aligned with the HR image, facilitating pixel alignment for loss computations. Additionally, the $Mim_{LR}$ retains the degradation and quality of the original LR image, allowing it to be replaced by the original LR during inference without additional manipulation. Our generation module is plug-and-play, integrating seamlessly with SR models in an end-to-end learning manner. We show that our model significantly improves the performance of existing SR models, from CNN to Transformer architectures, and from large models to efficient ones, outperforming state-of-the-art alignment attempts.

\noindent To conclude, the contribution of this paper is twofold:

\begin{itemize} 
    \item We address the misalignment issue in existing SR works and thoroughly revise the existing alignment processes by establishing comprehensive benchmarks on synthetic and realistic datasets.
    \item We propose a novel plug-and-play module that mimics the aligned LR to improve the SR learning stage, enhancing the robustness of SR models against misalignment without adding any learnable parameters during inference.
    \item Extensive comparisons across diverse SR models, ranging from CNNs to Transformers and from lightweight to heavy models, validate the effectiveness and generalization ability of our proposed method.
\end{itemize}

\section{Related Work}
\label{sec:related_work}
\noindent\textbf{Realistic Super Resolution}
Most research on Single Image Super Resolution (SISR) \cite{dong2014learning,lim2017enhanced,zhang2018residual,hui2019lightweight,zhang2022efficient,lu2022transformer,wang2023omni} relies on synthetic datasets with simple degradation models. To develop more robust SR models, researchers have been working on generating synthetic SR datasets with more complex degradation. These methods often involve degradation pipelines \cite{kohler2019toward,zhang2021designing,ji2020real} or parameterized degradation processes to create blur kernels. Degradation parameters can be derived through numerical methods \cite{shao2015simple,shao2019nonparametric} or optimized jointly using deep learning \cite{gu2019blind,cornillere2019blind,huang2020unfolding}. Additionally, some approaches leverage deep learning to generate training samples from pre-estimated degradation parameters \cite{bell2019blind,bulat2018learn,zhou2019kernel,xiao2020degradation}.

Another line of research tackles real-world super-resolution through domain adaptation \cite{yuan2018unsupervised,zhang2019multiple,kim2020unsupervised,maeda2020unpaired,prajapati2020unsupervised,fritsche2019frequency,lugmayr2019unsupervised,chen2020unsupervised}. In these works, real-world LR images and synthetic clean LR images are treated as different domains, with the goal of narrowing the gap between them. To make SR models more generalizable, self-learning-based methods have been employed to minimize the discrepancy between training and testing data \cite{zhao2020smore,shocher2018zero,emad2021dualsr,soh2020meta,park2020fast,zhang2022self}.

To address real-world SR in a fully supervised manner, several realistic SR datasets have been introduced to create paired image data. These datasets often try to capture LR and HR images of the same scene by varying the focal length of a zoom lens \cite{cai2019toward,zhang2019zoom,wei2020component,chen2019camera}. Other approaches utilize pixel binning \cite{kohler2019toward} and beam-splitter techniques \cite{joze2020imagepairs}. However, all of these datasets suffer from misalignment issues due to variations in the capture systems, optical setups, or imaging conditions. In our work, we focus on real-world SR using realistic SR datasets captured with zoom lenses.

\noindent\textbf{Alignment of Paired Images}
Misalignment is a common challenge in low-level vision. When capturing realistic datasets, multiple devices, varying optical systems, or changing capture settings are often required, leading to misalignment between the input and GT pairs. This problem is evident in Learned-Based ISP \cite{ignatov2020replacing,ignatov2021learned,shekhar2022transform,elezabi2024simple}, Realistic SR \cite{cai2019toward,zhang2019zoom,zhang2022self}, and image restoration \cite{feng2023generating}. A standard approach to mitigate misalignment is to apply global alignment techniques like affine transformations \cite{cai2019toward,ignatov2020replacing} or dense alignment using deep learning methods \cite{ignatov2021learned}.

However, global alignment alone is insufficient to achieve pixel-level correspondence between image pairs. To address this, the Contextual Loss (CXLoss) \cite{mechrez2018contextual} and the Contextual Bilateral Loss (CoBi Loss) \cite{mechrez2018contextual} were introduced for training on misaligned image pairs by computing the loss based on the most similar features between the compared pair. Differently, zhang et al. \cite{zhang2021learning} proposed an alignment that densely aligns the GT patch with a color-reconstructed version of the RAW input using optical flow

Other works generate a new pair that is better aligned. Zhang et al. \cite{zhang2022self} generates an auxiliary LR that is better aligned with the GT and used to train the restoration model. Feng et al. \cite{feng2023generating} instead produces a pseudo-GT better aligned with the input. They utilize a domain alignment module, for colors and degradations, and a geometric alignment module for structural alignment.

\section{Methodology}
\label{sec:methodology}
\begin{figure}[tb]
  \centering
  \includegraphics[trim={0 5.5mm 0 2.5mm},clip,width=\linewidth]{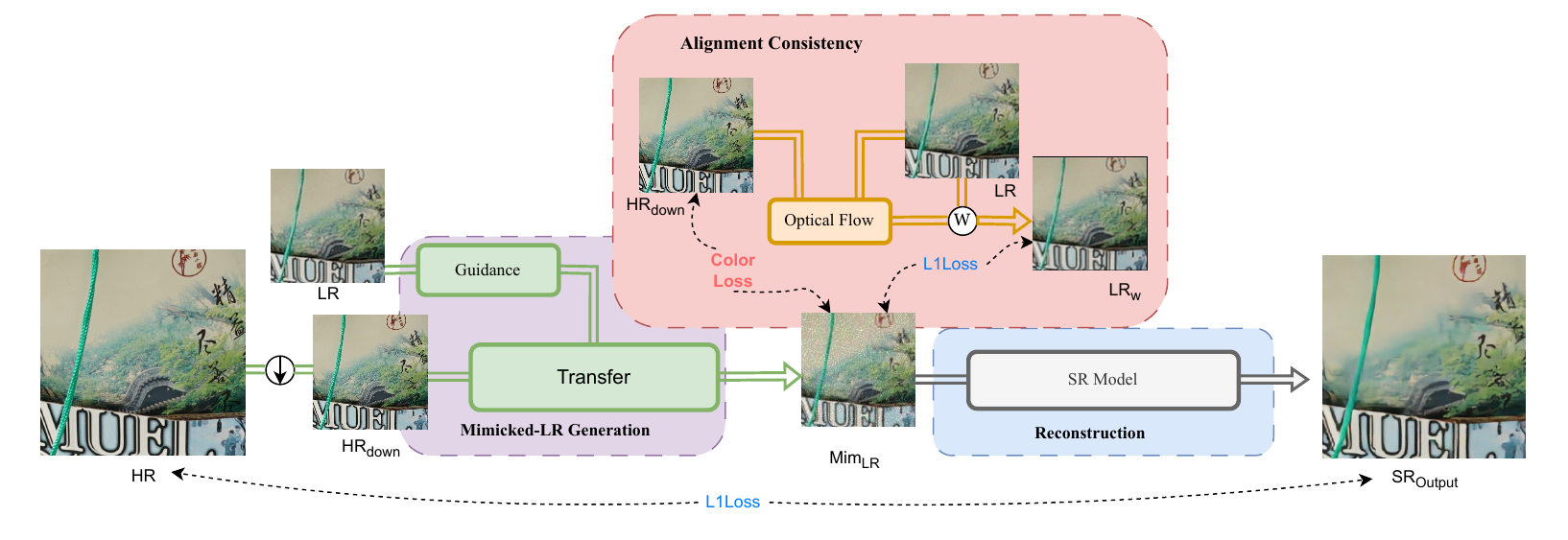}
  \caption{\textbf{Overview}. In the training stage, we employ a modification module for alignment purposes to create a Mimicked-LR input with improved consistency with the GT. Such module generates or mimics a new LR ($Mim_{LR}$) which (i) has the same degradations as the LR input image, (ii) is geometrically aligned with the HR, and (iii) is color consistent with the GT (same colors, brightness, etc). During inference, we remove the generation/alignment module and replace the Mimicked-LR with the normal LR input, allowing for a sharper output without color changes or distortion. }
  \label{fig:main}
\end{figure}

\noindent \textbf{Motivation:} We propose a module designed for seamless integration with any SR model to address alignment issues. This module generates a Mimicked-LR image that replaces the original LR image during training. The Mimicked-LR must meet three key criteria: First, it should be geometrically well-aligned with the GT HR image. Second, it must have the same color characteristics as the GT HR image to prevent model confusion and color artifacts. Third, the Mimicked-LR should accurately replicate the degradation characteristics of the original LR image, enabling a direct replacement during testing without performance loss. Hence, the model learns the SR task without being hindered by non-SR-related differences in the dataset. The overall pipeline is shown in Fig. \ref{fig:main}.

\subsection{Mimicked-LR Generation} 

To generate the Mimicked-LR, we design a straightforward yet effective architecture without any specialized layers. As depicted in Fig. \ref{fig:main}, we process the downscaled HR and LR images, transforming the downscaled HR to match the degradation of the LR image. The model relies on simple convolutions, ensuring efficient computation and minimal training cost, which allows for easy integration with any existing SR models without extra computational burden.

\begin{figure}[tb]
  \centering
  \includegraphics[trim={0 3.5mm 0 3.5mm},clip,width=.7\linewidth]{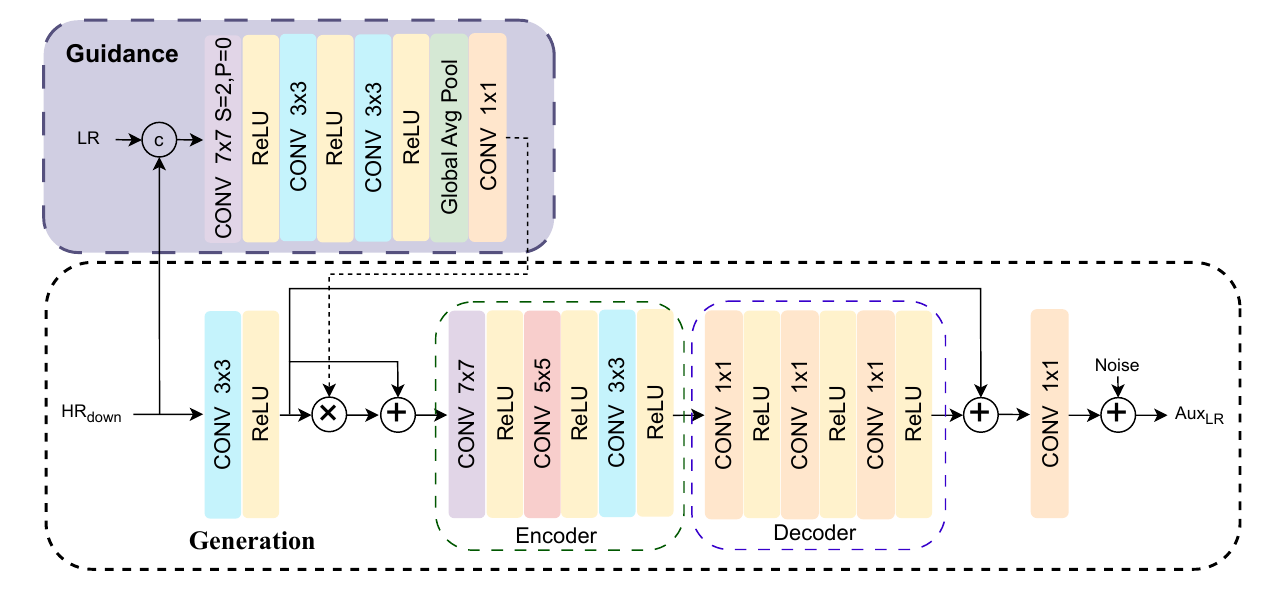}
  \caption{Illustration of the proposed LR Mimicking Architecture.}
  \label{fig:aux_generation}
\end{figure}

\noindent \textbf{Guidance Network:} Realistic degradation is non-uniform and can be scene-specific, varying for each image patch during training. To address this, we incorporate information from the LR image through a guidance network. As shown in Fig. \ref{fig:aux_generation}, the guidance network comprises a sequence of CNN layers that extract a guidance vector from the LR and downscaled HR images. We follow the shallow layers from ResNet, enhancing them with additional convolutions for deep feature extraction. The extracted features are then transformed into a guidance vector using global average pooling, capturing the overall style and domain information from the LR input. This vector further serves as an attention mechanism to guide the Mimicked-LR generation.

\noindent \textbf{Transfer Network:} The transfer network aims to transform the downscaled HR to match the LR degradation without introducing color or geometric shifts. Initially, we apply the guidance vector as a direct scale and shift to process the downscaled HR, projecting features from the HR domain into the target LR domain. The combined features are then processed through an encoder-decoder architecture for image generation.

Specifically, the encoder consists of convolutions with varying receptive fields to extract features in a coarse-to-fine manner, understanding global dependencies and better modeling the gap to align the two domains. The decoder reconstructs the degraded HR patch (Mimicked-LR) from the encoded features, using only $1\times 1$ convolution layers to avoid pixel shifting for better alignment. We incorporate conventional practices in low-level vision by adding skip connections and a final convolution layer. Additionally, we introduce noise at the end to simulate complex degradation, utilizing a mix of Gaussian and JPEG noise. The modification process maintains the same spatial size, preventing artifacts and shifts that typically result from downsampling and upsampling.

\subsection{Objective Functions} 
\textbf{Geometrical Alignment}: To ensure the geometrical alignment between the Mimicked-LR and GT, we use a downsampled version of the HR image (to the same size as the LR image) modified to mimic the characteristics of the LR. Inspired by \cite{zhang2021learning}, we adapt a misalignment loss using an optical flow network to densely align the compared image patches before applying the loss. 

Specifically, we calculate the optical flow between the downsampled HR patch $\widehat{HR}$ and the LR patch $LR$, then use the optical flow to align the LR patch $LR_{w}$. We compute the L1 loss between $LR_{w}$ and the output Mimicked-LR $Mim_{LR}$, ensuring geometrical alignment between $Mim_{LR}$ and GT $HR$ while retaining the characteristics of the LR. By limiting the optical flow alignment to the loss calculation, we avoid warping artifacts in the generated images.

Unlike previous works \cite{zhang2022self,sun2021learning}, our method modifies $\widehat{HR}$ to match the degradation of the LR rather than generating a new image, simplifying the task and avoiding generative artifacts. The misalignment loss is defined as follows:

\begin{align}
&LR_{w} = \mathcal{W}(LR, \psi), \, \, \, \,\,\,\,\,\, \psi = \mathcal{F}(\widehat{HR},LR)\\
&\mathcal{L}_{deg} = ||m\cdot(LR_{w}-Mim_{LR})||_1, \, \, \, \,\,\,\,\,\, 
Mim_{LR} = \mathcal{M}(\widehat{HR}, LR)\\
&m_i = \begin{cases}
    1,  [\omega(1, \psi)]_i \geq 1 - \epsilon \\
    0,  \quad otherwise
    \end{cases}
\end{align}

where $\mathcal{F}(.)$ and $\mathcal{W}(.)$ denote the optical flow and the warping operation, respectively. $m_i$ is the mask representing valid positions in the optical flow, and  $\mathcal{M}(.)$ is the Mimicked-LR creation module. $\mathcal{L}_{deg}$ represents the degradation loss.

\noindent \textbf{Color Alignment}: Ensuring color consistency between the input and GT during training is crucial to avoid model confusion and color artifacts. Inspired by \cite{elezabi2024simple}, we employ the Color Difference Network (CDNet) \cite{wang2023measuring} as an off-the-shelf color loss function. CDNet, trained on large-scale color difference datasets, achieves state-of-the-art performance and is fully differentiable, making it suitable for backpropagation. We formulate our color loss as follows:

\begin{equation}
    \mathcal{L}_{CD} = CD(Mim_{LR},\widehat{HR})
\end{equation}

\subsection{Learning Diagram}
\noindent\textbf{Training}: In the training stage, we integrate our Mimicking module with the SR model, replacing the conventional misaligned LR-HR pair with the Mimicked-LR-HR pair. We jointly train the Mimicking module and the SR model for optimization. To prevent the SR model from interfering with the Mimicking module's output, we detach the $Mim_{LR}$ image from the computation graph. This ensures that the SR model's backpropagation does not affect the weights of the Mimicking module, maintaining the desired characteristics of the $Mim_{LR}$ image and preventing it from being optimized to simplify the SR task. The complete loss function is described in the following equation:

\begin{align}
    &\mathcal{L}_{res} = ||SR_{Out}-HR||_1, \, \, \, \,\,\,\,\,\,  SR_{Out} = \mathcal{R}(Mim_{LR})\\
    &\mathcal{L}_{total} = \mathcal{L}_{res} + \mathcal{L}_{deg} + \lambda \mathcal{L}_{CD}
\end{align}
where $\mathcal{R}(.)$ is the SR Model. $\lambda$ is the balancing factor between the degradation loss $\mathcal{L}_{deg}$ and color difference loss $\mathcal{L}_{CD}$ for the mimicking module optimization. 

\noindent \textbf{Inference}: During inference, we simply remove our Mimicking module and replace $Mim_{LR}$ with the conventional $LR$. Our process ensures that $Mim_{LR}$ has the same degradation characteristics as the LR image, allowing this substitution to be made seamlessly without introducing artifacts or performance drop.


\section{Experiments}
\label{sec:results}
\subsection{Overall Evaluation Protocol}

\noindent This work aims to assess the robustness of the previous alignment process for SR models. To achieve this, we combine these alignment methods with widely used SR models and \textit{\textbf{retrain them on real-world datasets}} that inherently contain misalignment, while \textit{\textbf{evaluating}} their performance using \textit{\textbf{synthetic datasets}} with perfectly aligned LR-HR pairs. Notably, conventional SR metrics (PSNR/SSIM) cannot be reliably evaluated on the real-world dataset due to misalignment in the testing set; therefore, we compare the non-reference image quality metrics (NIQE \cite{mittal2012making}, NRQM \cite{ma2017learning}, PI \cite{blau20182018}).

\subsection{Datasets}
\label{data}

\noindent \textbf{Training:} Specifically, our experiments involve training on two distinct real-world datasets: RealSR \cite{cai2019toward} and SR-RAW \cite{zhang2019zoom}. 

\noindent\textbf{RealSR} \cite{cai2019toward} is a realistic dataset captured using DSLR cameras equipped with zoom lenses. LR and HR images were obtained by varying the focal length (28, 35, 50, and 105 mm) to explore different scaling factors. The dataset employs an image registration framework to achieve pixel-wise alignment. This process begins by cropping a central region from the HR image to mitigate severe distortions, serving as a reference for aligning images captured with other focal lengths. Optimization involves iteratively adjusting an affine transformation matrix and luminance parameters. Despite these efforts, residual misalignment remains evident, as depicted in Fig. \ref{fig:miss-alignment}. We focus our experiments on the $\times$4 scaling factor.

\noindent \textbf{SR-RAW} \cite{zhang2019zoom}, on the other hand, is captured using a similar process but includes a broader range of focal lengths (24–240 mm) and offers both RAW and RGB images. For our study, we utilize RGB images and focus on specific pairs (24/100, 35/150, and 50/240) to form a 4$\times$ scale dataset. The initial alignment process utilizes the Euclidean motion model \cite{evangelidis2008parametric}, resulting in more complex scenes and higher-resolution images compared to RealSR. However, this registration method also leads to image pairs with noticeable pixel-wise misalignment.

\noindent \textbf{Testing:} For comprehensive evaluation under full-reference conditions, we utilize synthetic SR benchmark datasets renowned for their absence of misalignment issues: Set5 \cite{bevilacqua2012low}, Set14 \cite{zeyde2012single}, BSD100 \cite{martin2001database}, Urban100 \cite{huang2015single}, Manga109 \cite{matsui2017sketch}, and DIV2KRK \cite{bell2019blind}, derived from the DIV2K dataset \cite{agustsson2017ntire} with complex degradation pipelines. Additionally, we incorporate non-reference image quality metrics to assess performance on the previously mentioned real-world datasets RealSR \cite{cai2019toward} and SR-RAW \cite{zhang2019zoom}, using their official training/testing splits.

\subsection{Implementation Details.}
For a consistent and fair comparison, we employ identical hyperparameters across all alignment processes tested on various SR models. We perform random cropping of patches sized 128$\times$ 128, augmented with random flipping and rotation. The reconstruction loss used for all models is L1 loss. Optimization is carried out using Adam optimizer \cite{kingma2014adam} for 200,000 iterations, with a batch size of 32 and an initial learning rate of 1e-3. To schedule the learning rate, we employ a Cosine Annealing LR scheduler \cite{loshchilov2016sgdr}. All experiments are implemented in the PyTorch framework on a single NVIDIA RTX 4090 GPU.

\subsection{SOTA Alignment Counterparts}
\label{training_processes}
For a comprehensive benchmark, we test 5 alignment processes including processes created for different tasks. In Tab. \ref{tab:ssraw}, \ref{tab:realsr}, \ref{tab:noref} Alignment Loss is the training process presented in \cite{zhang2021learning}. It densely aligns the gt patch with the input patch using optical flow. CXLoss is the contextual loss proposed in \cite{mechrez2018contextual} and the  CoBi Loss is Contextual Bilateral Loss presented in \cite{zhang2019zoom}. They compute loss on similar features between the compared images. Auxiliary Input uses the auxiliary-LR generation module from \cite{zhang2022self} to generate an auxiliary LR that replaces the LR during training. lastly, Pseudo GT is the alignment process proposed in \cite{feng2023generating}. They apply domain and geometrical alignment using two models to create a Pseudo GT that is used during training. We adapt these processes to work for the realistic SR task. 

Moreover, we evaluate the effectiveness of the tested alignment processes across four state-of-the-art (SOTA) SR models. Our selection includes two CNN-based models: RRDBNet \cite{wang2018esrgan} and SeeMore \cite{zamfir2024see}, as well as two transformer-based models: SwinIR \cite{liang2021swinir} and DAT \cite{chen2023dual}. These models differ not only in their architectural foundations but also in terms of model complexity and size. This diverse selection allows us to assess alignment techniques across a spectrum of SR methodologies.

\subsection{Full-Reference Quantitative Benchmark on Synthetic Data}

We employ synthetic benchmarks (Evaluation in Sec. \ref{data}) for quantitative full-reference evaluation to ensure no misalignment between the input and the ground truth (GT). Despite the different distributions of our training data (realistic SR datasets), which typically yield lower performance on synthetic datasets, they facilitate a valuable relative comparison between the alignment approaches. By evaluating the same SR model trained with different alignment processes, we aim to identify the optimal alignment strategy that produces sharper outputs without introducing color shifts. Such outputs typically achieve superior performance in synthetic benchmarks. Conversely, inadequate alignment processes during training lead to SR models that introduce geometric and color shifts, resulting in blurred outputs with noticeable color artifacts. These outputs typically perform poorly on synthetic benchmarks, underscoring the importance of effective misalignment correction strategies.

\begin{table}[tb]
  \caption{Comparison of the different alignment processes on the synthetic benchmark. All models are trained on \textbf{SR-RAW} dataset\cite{zhang2019zoom}. {\color{red}Best} and {\color{blue}second best} performances are highlighted. Our training processes archive the best performance in all tested methods. Please refer to Training Processes Evaluated Sec. \ref{training_processes} for information about training processes.}
  \label{tab:ssraw}
  \centering
  \resizebox{\linewidth}{!}{
  \begin{tabular}{l@{\qquad}l@{\qquad}ll@{\quad}ll@{\quad}ll@{\quad}ll@{\quad}ll@{\quad}ll}
    \toprule
    \multirow{2}{*}{\textbf{Methods}} & \multirow{2}{*}{\textbf{\thead{Training \\ Processes}}} & \multicolumn{2}{c}{\textbf{Set5}} & \multicolumn{2}{c}{\textbf{Set14}} & \multicolumn{2}{c}{\textbf{BSD100}} & \multicolumn{2}{c}{\textbf{Urban100}} & \multicolumn{2}{c}{\textbf{Manga109}}&\multicolumn{2}{c}{\textbf{DIV2KRK}}\\
    \cline{3-14}
    & & PSNR & SSIM & PSNR & SSIM & PSNR & SSIM & PSNR & SSIM & PSNR & SSIM & PSNR & SSIM\\
    \midrule
    \multirow{6}{*}{\textbf{RDDBNet\cite{wang2018esrgan}}} &
    Alignment Loss & 20.07 & \color{blue}0.7486 & 17.69 & \color{blue}0.6381 & 18.00 & \color{blue}0.6019 & 14.41 & 0.5755 & 17.33 & 0.7104& 15.46&0.5166\\
    {} &CXLoss & \color{blue}22.42 & 0.7017 & \color{blue}20.83 & 0.5902 & \color{blue}21.15 & 0.5446 & 18.85 & 0.5641 & 19.44 & 0.6939&19.69&\color{blue}0.5888\\
    {} &CoBi Loss & 22.02 & 0.5625 & 20.60 & 0.4650 & 20.49 & 0.4049 & \color{blue}19.35 & 0.4798 & 20.10 & 0.5895&\color{blue}21.02&0.4909\\
    {} &Auxiliary Input & 18.20 & 0.6996 & 16.88 & 0.5809 & 19.59 & 0.5601 & 14.24 & 0.4937 & 15.73 & 0.6460&15.10&0.5258\\
    {} &Pseudo GT & 22.11 & 0.7479 & 20.18 & 0.6338 & 20.19 & 0.5964 & 17.61 & \color{blue}0.5844 & \color{blue}20.05 & \color{blue}0.7196&17.66&0.5654\\
    {} & Ours & \color{red} 23.03 & \color{red}0.7599 & \color{red}22.73 & \color{red}0.6618 & \color{red}23.26 & \color{red}0.6356 & \color{red}19.99 & \color{red}0.6399 & \color{red}21.42 & \color{red}0.7811&\color{red}22.95&\color{red}0.6342\\
    
    \midrule
    \multirow{6}{*}{\textbf{SwinIR\cite{liang2021swinir}}} &
    Alignment Loss & 22.74 & \color{red}0.7779 & 22.26 & \color{blue}0.6676 & 22.04 & \color{blue}0.6337 & 19.91 & \color{blue}0.6350 & \color{blue}21.35 & \color{blue}0.7526&\color{blue}20.40&\color{blue}0.6125\\
    {} &CXLoss & 22.39 & 0.6144 & 21.05 & 0.5247 & 20.52 & 0.4743 & 19.27 & 0.5362 & 20.46 & 0.6435&18.93&0.2947\\
    {} &CoBi Loss & 21.18 & 0.5835 & 19.08 & 0.4693 & 17.86 & 0.3890 & 18.44 & 0.4874 & 19.44 & 0.5901&19.19&0.3143\\
    {} &Auxiliary Input & 21.01 & 0.7433 & 19.33 & 0.6272 & 21.09 & 0.5987 & 17.79 & 0.5865 & 19.94 & 0.7222&19.76&0.6076\\
    {} &Pseudo GT & \color{blue}22.96& 0.7453 & \color{blue}22.48 & 0.6406 & \color{blue}23.10 & 0.6091 & \color{blue}20.45 & 0.6011 & 20.88 & 0.7247&19.58&0.5945\\
    {} & Ours & \color{red}23.73 & \color{blue}0.7748 & \color{red}23.26 & \color{red}0.6725 & \color{red}24.04 & \color{red}0.6434 & \color{red}20.98 & \color{red}0.6555 & \color{red}22.90 & \color{red}0.7947&\color{red}23.41&\color{red}0.6464\\

    \midrule
    \multirow{6}{*}{\textbf{DAT\cite{chen2023dual}}} &
    Alignment Loss & 21.99 & \color{blue}0.7632 & \color{blue}22.76 & \color{blue}0.6644 & 22.40 & \color{blue}0.6257 & \color{blue}20.73 & \color{blue}0.6361 & 21.41 & \color{blue}0.7478&21.46&0.6132\\
    {} &CXLoss & 22.02 & 0.5625 & 20.60 & 0.4650 & 20.49 & 0.4049 & 19.35 & 0.4798 & 20.10 & 0.5895&19.07&0.2772\\
    {} &CoBi Loss & 22.34 & 0.5849 & 20.53 & 0.4796 & 19.09 & 0.3925 & 19.21 & 0.4764 & 20.50 & 0.5955&19.92&0.2873\\
    {} &Auxiliary Input & 22.67 & 0.7496 & 19.79 & 0.6226 & 21.65 & 0.5937 & 18.89 & 0.5830 & 20.68 & 0.7141&\color{blue}21.49&\color{blue}0.6164\\
    {} &Pseudo GT & \color{red}23.70 & 0.7399 & \color{red}23.34 & 0.6442 & \color{blue}23.37 & 0.6029 & \color{red}20.92 & 0.6033 & \color{blue}21.53 & 0.7294&20.63&0.6046\\
    {} & Ours & \color{blue}22.87 & \color{red}0.7652 & 22.40 & \color{red}0.6690 & \color{red}23.39 & \color{red}0.6425 & 20.13 & \color{red}0.6497 & \color{red}22.48 & \color{red}0.7925&\color{red}23.61&\color{red}0.6533\\

    \midrule
    \multirow{6}{*}{\textbf{SeeMore\cite{zamfir2024see}}} &
    Alignment Loss & 21.85 & \color{red}0.7718 & 20.90 & \color{blue}0.6668 & 20.69 & \color{blue}0.6246 & 18.45 & \color{blue}0.6207 & \color{blue}20.23 & \color{blue}0.7439&17.84&\color{blue}0.5843\\
    {} &CXLoss & 21.02 & 0.5355 & 19.84 & 0.4397 & 18.68 & 0.3766 & 17.46 & 0.4297 & 19.28 & 0.5394&16.51&0.2488\\
    {} &CoBi Loss & 20.85 & 0.5751 & 18.55 & 0.4614 & 17.64 & 0.3706 & 17.63 & 0.4600 & 18.67 & 0.5738&\color{blue}18.02&0.2721\\
    {} &Auxiliary Input & 18.49 & 0.7087 & 17.25 & 0.6035 & 18.80 & 0.5851 & 14.901 & 0.5526 & 17.94 & 0.6866&17.50&0.5813\\
    {} &Pseudo GT & \color{blue}22.45 & 0.7429 & \color{blue}21.54 & 0.6389 & \color{blue}21.76 & 0.6036 & \color{blue}19.20 & 0.5923 & 19.40 & 0.7159&17.60&0.5498\\
    {} & Ours & \color{red}23.90 & \color{blue}0.7626 & \color{red}23.28 & \color{red}0.6669 & \color{red}23.56 & \color{red}0.6348 & \color{red}21.00 & \color{red}0.6410 & \color{red}22.35 & \color{red}0.7797&\color{red}23.33&\color{red}0.6398\\
    
  \bottomrule
  \end{tabular}
  }
\end{table}

\begin{table}[tb]
  \caption{Comparison of the different alignment processes on the synthetic benchmark. All models are trained on \textbf{RealSR} dataset\cite{cai2019toward}. {\color{red}Best} and {\color{blue}second best} performances are highlighted. Our training processes archive the best performance in all tested methods. Please refer to Training Processes Evaluated Sec. \ref{training_processes} for information about training processes.}
  \label{tab:realsr}
  \centering
  \resizebox{\linewidth}{!}{
  \begin{tabular}{l@{\qquad}l@{\qquad}ll@{\quad}ll@{\quad}ll@{\quad}ll@{\quad}ll@{\quad}ll}
    \toprule
    \multirow{2}{*}{\textbf{Methods}} & \multirow{2}{*}{\textbf{\thead{Training \\ Processes}}} & \multicolumn{2}{c}{\textbf{Set5}} & \multicolumn{2}{c}{\textbf{Set14}} & \multicolumn{2}{c}{\textbf{BSD100}} & \multicolumn{2}{c}{\textbf{Urban100}} & \multicolumn{2}{c}{\textbf{Manga109}}&\multicolumn{2}{c}{\textbf{DIV2KRK}}\\
    \cline{3-14}
    & & PSNR & SSIM & PSNR & SSIM & PSNR & SSIM & PSNR & SSIM & PSNR & SSIM & PSNR & SSIM\\
    \midrule
    \multirow{7}{*}{\textbf{RDDBNet\cite{wang2018esrgan}}} &
    L1Loss & 22.40 & 0.6788 & 21.40 & 0.6025 & 21.067 & 0.5641 & 19.24 & 0.5643 & \color{blue}21.02 & 0.7079&\color{blue}23.89&0.6657\\
    {} & Alignment Loss & 22.48 & 0.6900 & 21.65 & 0.6078 & 21.28 & 0.5700 & \color{blue}19.61 & 0.5751 & 20.95 & 0.7266&\color{blue}23.89&0.6626\\
    {} &CXLoss & 20.76 & 0.5948 & 19.45 & 0.5141 & 18.88 & 0.4584 & 17.86 & 0.4994 & 19.66 & 0.6507&23.48&0.6268\\
    {} &CoBi Loss & 20.51 & 0.6102 & 19.11 & 0.5234 & 18.09 & 0.4570 & 17.30 & 0.5054 & 19.91 & 0.6764&23.42&0.6418\\
    {} &Auxiliary Input & \color{blue}22.94 & 0.6796 & \color{blue}22.05 & 0.5853 & \color{blue}22.59 & 0.5574 & 19.51 & 0.5206 & 20.49 & 0.6786&22.72&0.6302\\
    {} &Pseudo GT& 22.86 & \color{blue}0.7014 & 21.38 & \color{blue}0.6080 & 21.04 & \color{blue}0.5707 & 18.94 & 0.5625 & 20.74 & 0.7139&\color{red}23.98&\color{red}0.6703\\
    {} & Ours & \color{red}24.62 & \color{red}0.7587 & \color{red}23.45 & \color{red}0.6515 & \color{red}23.75 & \color{red}0.6289 & \color{red}21.16 & \color{red}0.6239 & \color{red}22.76 & \color{red}0.7815&23.56&\color{blue}0.6679\\

    \midrule
    \multirow{7}{*}{\textbf{SwinIR\cite{liang2021swinir}}} &
    L1Loss & 23.51 & 0.7153 & 21.85 & 0.6200 & 21.66 & 0.5883 & 19.77 & 0.5901 & 21.95 & 0.7476&\color{blue}24.10&\color{blue}0.6732\\
    {} & Alignment Loss & 23.92 & 0.7188 & 22.53 & 0.6342 & 22.08 & 0.5930 & \color{blue}20.61 & \color{blue}0.6126 & \color{blue}22.84 & \color{blue}0.7673&\color{blue}24.10&0.6715\\
    {} &CXLoss & 21.56 & 0.6043 & 20.40 & 0.5354 & 19.67 & 0.4808 & 19.14 & 0.5451 & 21.16 & 0.6917&22.49&0.4946\\
    {} &CoBi Loss & 20.65 & 0.5791 & 19.57 & 0.5155 & 18.43 & 0.4528 & 18.39 & 0.5223 & 20.71 & 0.6839&22.37&0.5285\\
    {} &Auxiliary Input & 23.96 & 0.7035 & \color{blue}22.61 & 0.6000 & \color{blue}23.02 & 0.5706 & 19.97 & 0.5413 & 21.19 & 0.7053&23.79&0.6501\\
    {} &Pseudo GT& \color{blue}24.11 & \color{blue}0.7397 & 22.52 & \color{blue}0.6491 & 22.09 & \color{blue}0.6093 & 20.26 & 0.6125 & 22.29 & 0.7643&24.01&0.6720\\
    {} & Ours & \color{red}24.69 & \color{red}0.7647 & \color{red}23.86 & \color{red}0.6680 & \color{red}23.83 & \color{red}0.6357 & \color{red}21.50 & \color{red}0.6462 & \color{red}23.81 & \color{red}0.7992&\color{red}24.27&\color{red}0.6797\\

    \midrule
    \multirow{7}{*}{\textbf{DAT\cite{chen2023dual}}} &
    L1Loss & 23.05 & 0.6950 & 21.70 & 0.6058 & 21.12 & 0.5633 & 19.42 & 0.5621 & 21.46 & 0.7246&24.14&0.6738\\
    {} & Alignment Loss & \color{blue}24.04 & \color{blue}0.7280 & 22.50 & \color{blue}0.6324 & 22.32 & \color{blue}0.6015 & \color{blue}20.46 & \color{blue}0.5956 & \color{blue}22.21 & \color{blue}0.7490&24.13&0.6726\\
    {} &CXLoss & 21.31 & 0.5816 & 20.60 & 0.5268 & 20.20 & 0.4937 & 18.98 & 0.5251 & 20.82 & 0.6657&22.59&0.5179\\
    {} &CoBi Loss & 20.25 & 0.5667 & 19.43 & 0.4982 & 18.61 & 0.4442 & 18.19 & 0.4980 & 20.37 & 0.6542&22.19&0.5035\\
    {} &Auxiliary Input & 24.10 & 0.7193 & \color{blue}22.83 & 0.6138 & \color{blue}23.12 & 0.5858 & 20.29 & 0.5684 & 21.55 & 0.7222&23.95&0.6612\\
    {} &Pseudo GT& 23.67 & 0.7112 & 21.99 & 0.6246 & 21.44 & 0.5812 & 19.68 & 0.5859 & 21.77 & 0.7439&\color{blue}24.13&\color{blue}0.6766\\
    {} & Ours & \color{red}24.82 & \color{red}0.7710 & \color{red}24.05 & \color{red}0.6739 & \color{red}23.86 & \color{red}0.6433 & \color{red}21.80 & \color{red}0.6592 & \color{red}23.77 & \color{red}0.8045&\color{red}24.35&\color{red}0.6885\\

    \midrule
    \multirow{7}{*}{\textbf{SeeMore\cite{zamfir2024see}}} &
    L1Loss & 23.45 & 0.7047 & 22.24 & 0.6275 & 21.73 & 0.5838 & 20.01 & 0.5849 & 21.83 & 0.7305&23.80&0.6631\\
    {} & Alignment Loss & 23.75 & 0.7067 & 22.76 & \color{blue}0.6424 & 22.64 & \color{red}0.6216 & \color{blue}20.45 & 0.5998 & 21.56 & 0.7290&23.06&0.6412\\
    {} &CXLoss & 21.96 & 0.5923 & 21.33 & 0.5410 & 20.92 & 0.4965 & 19.46 & 0.5098 & 21.68 & 0.6697&22.00&0.4789\\
    {} &CoBi Loss & 21.38 & 0.6052 & 20.35 & 0.5325 & 19.83 & 0.4893 & 18.70 & 0.5231 & 20.48 & 0.6535&22.34&0.5259\\
    {} &Auxiliary Input & 24.12 & 0.7201 & \color{blue}22.87 & 0.6108 & \color{red}23.35 & 0.5860 & 20.16 & 0.5584 & 21.43 & 0.7084&23.22&0.6464\\
    {} &Pseudo GT& \color{red}24.39 & \color{blue}0.7354 & 22.65 & \color{red}0.6478 & 22.32 & 0.6139 & \color{blue}20.45 & \color{blue}0.6074 & \color{blue}21.87 & \color{blue}0.7346&\color{blue}23.64&\color{blue}0.6590\\
    {} & Ours & \color{blue}24.15 & \color{red}0.7483 & \color{red}23.08 & \color{red}0.6478 & \color{blue}23.17 & \color{blue}0.6194 & \color{red}20.86 & \color{red}0.6237 & \color{red}22.88 & \color{red}0.7813&\color{red}24.14&\color{red}0.6773\\
    
  \bottomrule
  \end{tabular}
  }
\end{table}

\noindent \textbf{Training with SR-RAW Dataset:} The results of our models trained with different alignment processes on the SR-RAW dataset \cite{zhang2019zoom} and evaluated on synthetic data are presented in Tab. \ref{tab:ssraw}. Due to significant misalignment in the dataset, we omitted evaluation based on L1 loss, which resulted in severely blurred outputs. Across all tested methods at a 4$\times$ scale SR, our alignment process consistently achieves superior performance compared to other alignment methods. This performance advantage holds across various SR models, demonstrating the robustness and consistency of our approach regardless of the specific model architecture. In contrast, alternative alignment methods exhibit varying performance across different SR models, indicating disparities in effectiveness. Moreover, our method demonstrates balanced performance in both PSNR and SSIM metrics, unlike other methods that may prioritize one metric over the other due to the sensitivity of PSNR to blurred outputs. This underscores again the efficacy of our approach in achieving high-quality outputs.

Our approach leverages a simple module that is jointly trained with the SR model, eliminating the need for additional separate models (e.g., Pseudo GT) or complex calculations (e.g., CX Loss, CoBi Loss). Unlike alignment loss methods applied during the reconstruction stage, we integrate alignment considerations into the LR Mimicking stage, enhancing efficiency and performance. Furthermore, our method exhibits robustness compared to limited alignment modules used in the Auxiliary Input approach \cite{zhang2022self}. 

\noindent \textbf{Training with RealSR Dataset:} We can notice the consistent pattern on the models trained with RealSR dataset \cite{cai2019toward} as shown in Tab. \ref{tab:realsr}. Despite RealSR's relatively good alignment compared to the SR-RAW dataset \cite{zhang2019zoom}, we observe notable improvements when employing alignment processes. Even minor misalignment present in the dataset significantly impacts model performance, particularly evident in the poorer results obtained with L1 loss. This underscores the sensitivity of SR models to alignment issues, highlighting the necessity and effectiveness of robust alignment strategies for enhancing performance across diverse datasets.

\subsection{Realistic Data Benchmark}

\begin{table}[tb]
  \caption{Evaluation with Realistic SR Datasets on No Reference Image Quality Metrics. Models are trained and evaluated on the same dataset. {\color{red}Best} and {\color{blue}second best} performances are highlighted.}
  \label{tab:noref}
  \centering
  \resizebox{.7\linewidth}{!}{
  \begin{tabular}{l@{\qquad}l@{\qquad}lll@{\quad}lll@{\quad}}
    \toprule
    \multirow{2}{*}{\textbf{Methods}} & \multirow{2}{*}{\textbf{\thead{Training \\ Processes}}} & \multicolumn{3}{c}{\textbf{SR-RAW}} & \multicolumn{2}{c}{\textbf{RealSR}}\\
    \cline{3-8}
    & & NIQE$\downarrow$ & NRQM$\uparrow$ & PI$\downarrow$ & NIQE$\downarrow$ & NRQM$\uparrow$ & PI$\downarrow$ \\
    \midrule
    \multirow{7}{*}{\textbf{RDDBNet\cite{wang2018esrgan}}} &
    Synthetic Data & 7.42&3.24&7.19&8.74&2.92&7.99\\
    {} &Alignment Loss & 7.88&3.19&7.47&7.50&3.23&7.17\\
    {} &CXLoss & \color{red}6.16&\color{blue}{4.95}&6.84&7.61&\color{blue}{4.42}&\color{red}6.53\\
    {} &CoBi Loss & 8.96&\color{red}{5.22}&\color{red}5.67&\color{blue}7.41&\color{red}{4.46}&\color{red}6.53\\
    {} &Auxiliary Input & 7.02&3.09&7.52&7.63&3.09&7.35\\
    {} &Pseudo GT & 8.18&2.81&7.78&7.53&3.26&7.19\\
    {} & Ours & \color{blue}6.67&4.11&\color{blue}6.40&\color{red}7.07&3.47&\color{blue}6.79\\

    \midrule
    \multirow{7}{*}{\textbf{SwinIR\cite{liang2021swinir}}} &
    Synthetic Data &7.41&3.25&7.18&8.72&2.93&7.97\\
    {} &Alignment Loss & 7.74&3.13& 7.42& 7.49&3.17&7.19\\
    {} &CXLoss & 26.61&\color{red}{6.35}&14.90&16.189&\color{red}{4.97}&10.39\\
    {} &CoBi Loss & 27.06&\color{blue}{5.68}&15.28&17.14&\color{blue}{4.92}&10.80\\
    {} &Auxiliary Input & 7.39 &3.37&\color{blue}7.13&7.65&3.14&7.33\\
    {} &Pseudo GT & 8.45&2.86&7.88&\color{blue}7.40&3.21&\color{blue}7.15 \\
    {} & Ours & \color{red}6.94&3.73&\color{red}6.71&\color{red}6.94&3.39&\color{red}6.78\\

    \midrule
    \multirow{7}{*}{\textbf{DAT\cite{chen2023dual}}} &
    Synthetic Data &\color{blue}7.43&3.23&7.20&8.71&2.94&7.96\\
    {} &Alignment Loss & 7.96 &3.15&7.51&7.59&3.16&7.26\\
    {} &CXLoss & 32.84&\color{red}{6.15}&17.91&15.98&\color{blue}{4.95}&10.23\\
    {} &CoBi Loss & 24.18&\color{blue}{5.93}&13.73&17.62&\color{red}{5.08}&11.07\\
    {} &Auxiliary Input & 7.48&3.34&\color{blue}7.19&7.56&3.20&7.27\\
    {} &Pseudo GT & 8.69&2.90&7.97&\color{blue}7.49&3.24&\color{blue}7.18\\
    {} & Ours &\color{red}6.93&3.93&\color{red}6.58&\color{red}6.98&3.42&\color{red}6.78\\

    \midrule
    \multirow{7}{*}{\textbf{SeeMore\cite{zamfir2024see}}} &
    Synthetic Data &\color{blue}7.25&3.29&\color{blue}7.09&8.64&2.92&7.93\\
    {} &Alignment Loss &7.73&3.12&7.40&7.55&3.16&7.23\\
    {} &CXLoss & 27.23&\color{red}{6.43}&15.10&19.56&\color{red}{4.97}&11.96\\
    {} &CoBi Loss & 25.18&\color{blue}{5.98}&14.25&15.29&\color{blue}{4.92}&9.91\\
    {} &Auxiliary Input & 7.37&3.24&7.20&\color{blue}7.51&3.14&7.23\\
    {} &Pseudo GT & 8.50&2.78&7.93&\color{blue}7.51&3.20&\color{blue}7.17\\
    {} & Ours & \color{red}6.91&3.74&\color{red}6.68&\color{red}6.84&3.51&\color{red}6.65\\
    
  \bottomrule
  \end{tabular}
  }
\end{table}

Since our focus is on training with realistic SR datasets, evaluating alignment processes on these datasets' images is crucial. Due to the presence of misalignment issues in both training and validation datasets, traditional full-reference evaluation methods like PSNR and SSIM are not suitable. Therefore, we employed No Reference Evaluation methods for a more accurate comparison.

As observed from the results in Tab. \ref{tab:noref}, similar to the synthetic benchmark, our alignment process achieves superior performance by attaining the highest scores across the NIQE and PI, underscoring the high quality of our process outputs. Furthermore, our method demonstrates remarkable consistency in performance across different SR models. This improvement is primarily attributed to the consistent output quality and minimal color distortions, aspects that are effectively evaluated by no-reference metrics.

Note that we also include the no-reference metric, NRQM, in our evaluation. This metric is less robust to artifacts. For instance, methods using CXLoss and CoBi Loss may achieve higher NRQM scores but typically perform worse on NIQE and PI metrics. Please refer to the qualitative results in Sec. \ref{qual_res} for the visual artifacts when training with CXLoss and CoBi.  Despite this, our method consistently outperforms all other methods in terms of NRQM score, indicating our effectiveness compared to alternative approaches.

\subsection{Qualitative Results}
\label{qual_res}

\begin{figure}[tb]
  \centering
    \includegraphics[trim={0 2.5mm 0 2mm},clip,width=\textwidth]{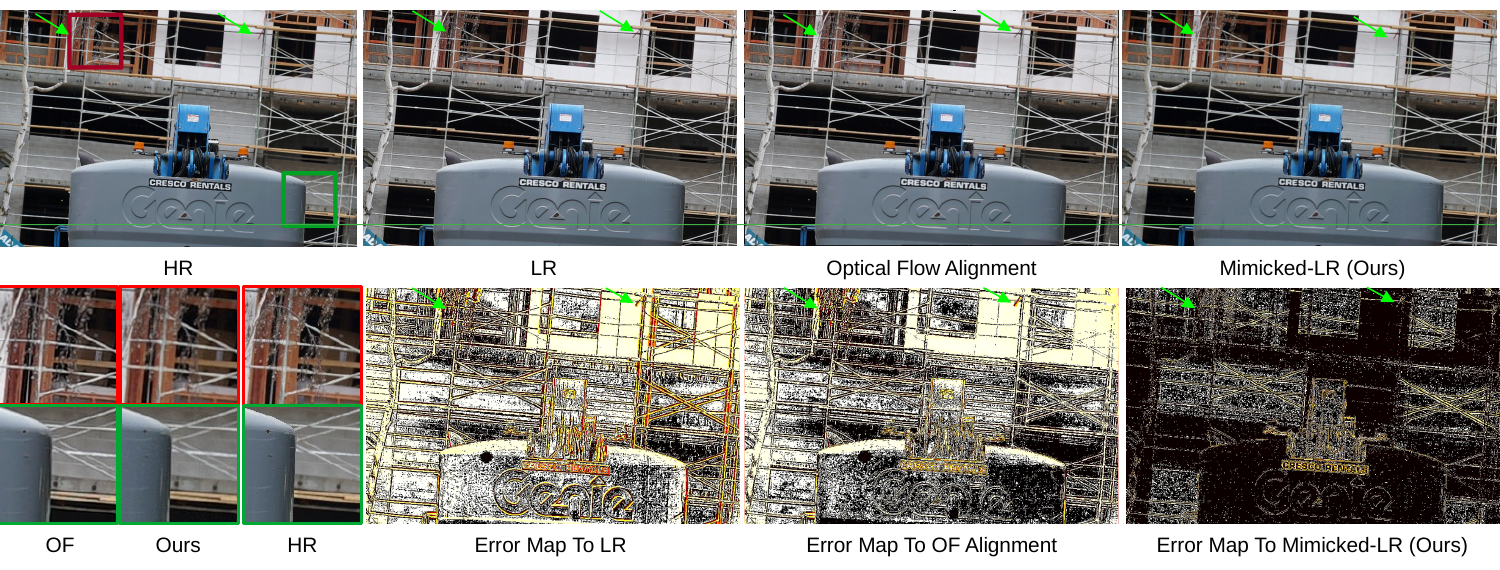}
  \caption{Visualization of our Mimicking alignment quality. We compare to dense alignment using Optical Flow. The error maps illustrate that our method produces outputs better aligned with the HR image.}
  \label{fig:ali-out}
\end{figure}

\begin{figure}[tb]
  \centering
    \includegraphics[trim={0 4.5mm 0 5.5mm},clip,width=\textwidth]{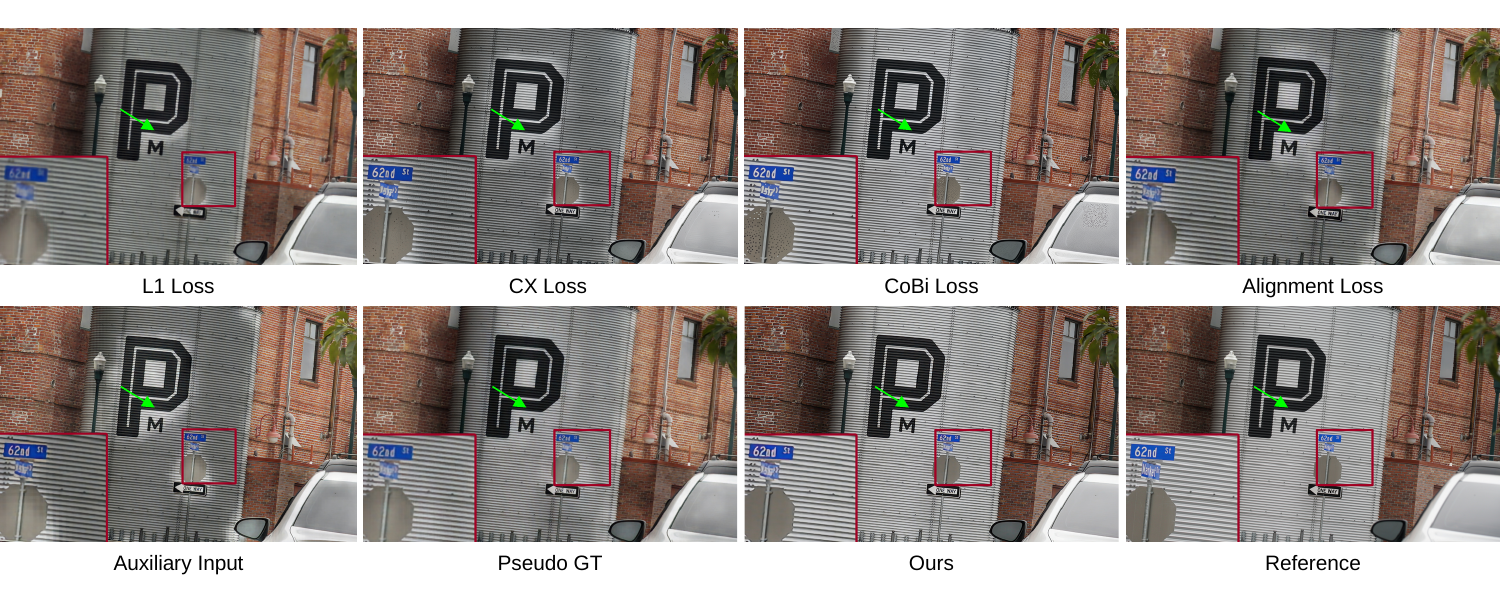}
  \caption{Qualitative Results of SwinIR\cite{liang2021swinir} Model Trained on SR-RAW Dataset}
  \label{fig:swinir}
\end{figure}

\begin{figure}[tb]
  \centering
    \includegraphics[trim={0 4.5mm 0 5.5mm},clip,width=\textwidth]{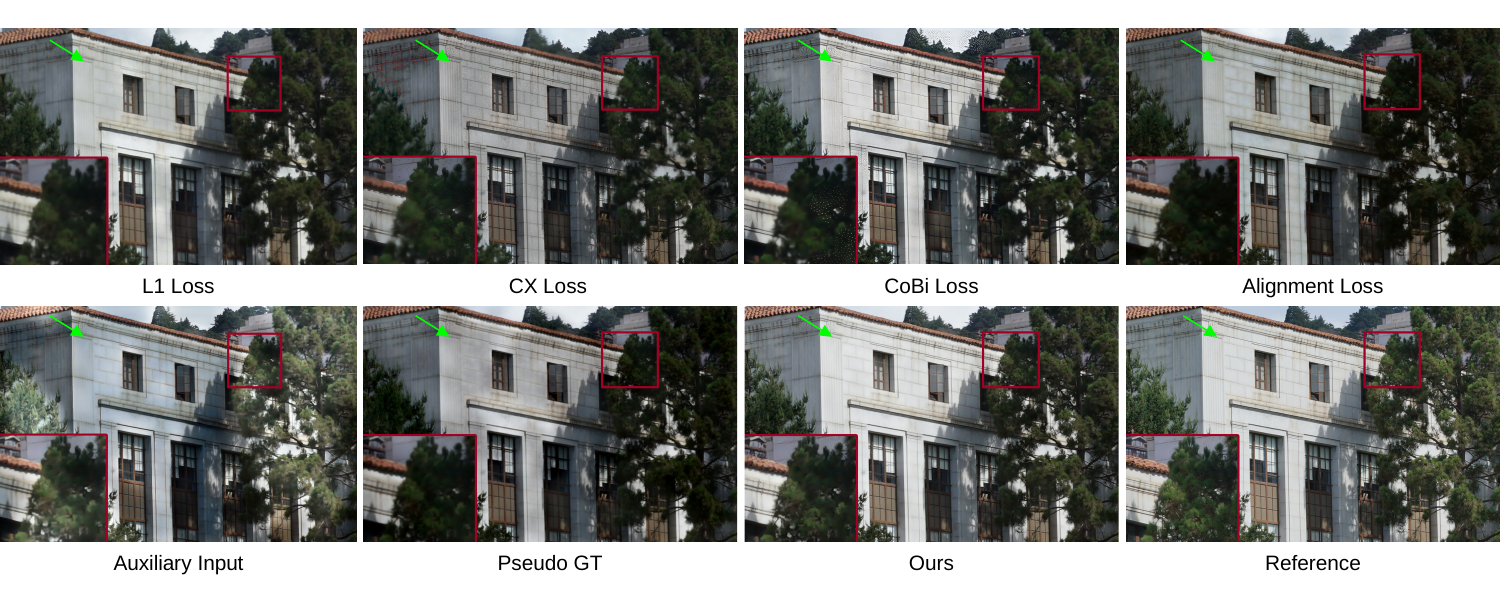}
  \caption{Qualitative Results of SeeMore\cite{zamfir2024see} Model Trained on SR-RAW Dataset }
  \label{fig:seemore}
\end{figure}

\noindent \textbf{Alignment Output:} First, we show the effectiveness of our alignment process in producing a better-aligned input. In Fig. \ref{fig:ali-out} we show the improved alignment compared to the SOTA counterpart based on optical flow (OF). Quantitatively, compared to the OF alignment, our method reduces the per-pixel error by  69.36\% and 9.77\% on SR-Raw (less aligned) and RealSR (better aligned), respectively. Additionally, we can notice the improved color alignment in the error map of the uniform areas which is not addressed by the OF method. Likewise, we can appreciate our method's ability to capture the LR image degradation producing a Mimicked LR image with similar characteristics. 

\noindent \textbf{SR Output:} For qualitative evaluation, we present the outputs of various SR models trained using different alignment processes.  
 In Fig. \ref{fig:swinir}, we showcase the results of the SwinIR \cite{liang2021swinir} and Fig. \ref{fig:seemore} of the SeeMoRe \cite{zamfir2024see} model trained on the SR-RAW dataset under different alignment methods. Our proposed alignment process consistently yields the best outcomes, producing sharp images devoid of color artifacts while preserving structural fidelity. In contrast, other alignment methods exhibit noticeable color misalignment issues, resulting in visible artifacts in their outputs. In Fig. \ref{fig:swinir} following the green arrow we can see severe color artifacts in the other alignment processes. Additionally, in the zoomed area we can see our method producing sharper details with better texture. 
 
 Similarly, in Fig. \ref{fig:seemore}, we can see severe color artifacts in the building with bad color reproduction. In the zoomed area we can see our method producing better shadows, better details, and texture. However, the Pseudo GT method mitigates color artifacts to some extent by considering dataset domain disparities, albeit with reduced robustness due to error accumulation from separately trained models. Models trained with CXLoss and CoBi Loss exhibit cartoon artifacts, as these methods compare features at the expense of fine image details.

 \begin{figure}[tb]
  \centering
    \includegraphics[trim={0 4.5mm 0 5.5mm},clip,width=\textwidth]{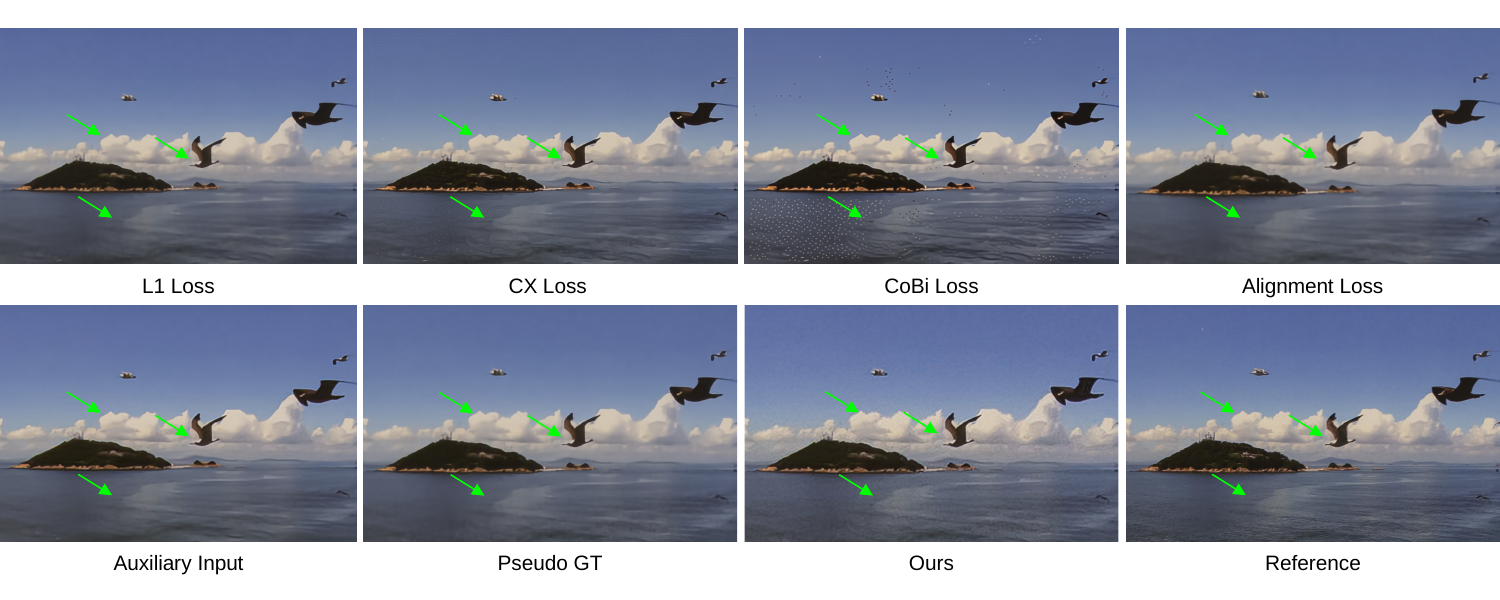}
    \label{fig:teaser_syns}
  \caption{Qualitative Results of DAT\cite{chen2023dual} Model Trained on RealSR Dataset. }
  \label{fig:dat}
\end{figure}

Additionally, we show the output of the different alignment processes on the RealSR dataset in Fig. \ref{fig:dat}. The performance improvement achieved through our method is evident. Our process yields sharper outputs with more details compared to the blurred and washed-out results from other methods. Following the green arrows, we observe enhanced textures and details in the clouds and water. These visual results, combined with the previously presented quantitative data, validate the effectiveness of our model in handling color and geometric misalignment in realistic SR datasets. Furthermore, our approach proves robust across different SR models and datasets, regardless of varying levels of misalignment.

\section{Conclusion}

We propose a novel training process designed to effectively train SR models on realistic SR datasets. Our approach involves a simple module that creates an image mimicking the LR image while maintaining alignment with the HR image. This module is plug-and-play, allowing it to be integrated with any SR model without additional modifications. Moreover, we introduce an extensive benchmark for various alignment processes across different datasets and SR models. Our training process consistently outperforms all other tested alignment methods. The superior visual quality of our method is evident when compared to other alignment processes, showcasing sharper, more detailed outputs without color changes or distortions. Our extensive evaluation demonstrates the effectiveness and generality of our method across different datasets and models. We believe our work lays the foundation for a new approach to addressing misalignment issues in fully supervised real-world datasets, potentially leading to significant advancements in the field of Super-Resolution.

\noindent \textbf{Acknowledgments:}  This work was supported by The Alexander von Humboldt Foundation.



\bibliographystyle{splncs04}
\bibliography{main}

\end{document}